\documentclass{article} 
\usepackage{iclr2027_conference,times}

\usepackage{amsmath,amsfonts,bm}

\def\eqref#1{equation~\ref{#1}}

\def\1{\bm{1}}

\DeclareMathAlphabet{\mathsfit}{\encodingdefault}{\sfdefault}{m}{sl}
\SetMathAlphabet{\mathsfit}{bold}{\encodingdefault}{\sfdefault}{bx}{n}

\usepackage{bbm}
\usepackage{caption}
\usepackage{wrapfig}
\usepackage{hyperref}
\usepackage{url}
\usepackage{lipsum}
\usepackage{graphicx}
\usepackage{booktabs}
\usepackage[table]{xcolor}
\usepackage{array}
\usepackage{threeparttable}
\usepackage{makecell}
\usepackage{rotating}
\usepackage{multirow}
\usepackage{colortbl}
\usepackage{footmisc}
\definecolor{barriercolor}{RGB}{255, 61, 98}             
\definecolor{bicyclecolor}{RGB}{0, 217, 128}             
\definecolor{buscolor}{RGB}{255, 230, 0}                 
\definecolor{carcolor}{RGB}{22, 140, 255}                
\definecolor{construction_vehiclecolor}{RGB}{255, 138, 0} 
\definecolor{motorcyclecolor}{RGB}{182, 44, 255}         
\definecolor{pedestriancolor}{RGB}{255, 61, 187}          
\definecolor{traffic_conecolor}{RGB}{255, 99, 63}        
\definecolor{trailercolor}{RGB}{0, 207, 232}             
\definecolor{truckcolor}{RGB}{255, 79, 79}               
\definecolor{E5CCFF}{HTML}{E5CCFF} 
\definecolor{CCCCFF}{HTML}{CCCCFF} 
\definecolor{driveable_surfacecolor}{RGB}{85, 214, 255}  
\definecolor{other_flatcolor}{RGB}{198, 108, 255}        
\definecolor{sidewalkcolor}{RGB}{217, 224, 230}          
\definecolor{terraincolor}{RGB}{182, 234, 56}            
\definecolor{manmadecolor}{RGB}{154, 168, 181}           
\definecolor{vegetationcolor}{RGB}{39, 232, 111}         
\definecolor{settingfull}{HTML}{F7FBFF}          
\definecolor{settingfront}{HTML}{edf6fc}         
\definecolor{settingfrontthree}{HTML}{e3f0f8}    
\definecolor{settingallcam}{HTML}{d9ebf5}        
\definecolor{settinglidar}{HTML}{cfe5f2}         

\definecolor{settingall}{HTML}{FFF7FA}       
\definecolor{settingallcamkt}{HTML}{F5DCE7}  
\definecolor{settinglidarkt}{HTML}{F1C9DA}   

\definecolor{tablegray}{HTML}{9A9A9A}
\definecolor{emptycolor}{RGB}{0, 0, 0}
\definecolor{carcolorkt}{RGB}{245, 150, 100}
\definecolor{bicyclecolorkt}{RGB}{245, 230, 100}
\definecolor{motorcyclecolorkt}{RGB}{150, 60, 30}
\definecolor{truckcolorkt}{RGB}{180, 30, 80}
\definecolor{other-vehiclecolorkt}{RGB}{255, 0, 0}
\definecolor{personcolor}{RGB}{30, 30, 255}
\definecolor{bicyclistcolor}{RGB}{200, 40, 255}
\definecolor{motorcyclistcolor}{RGB}{90, 30, 150}
\definecolor{roadcolor}{RGB}{255, 0, 255}
\definecolor{parkingcolor}{RGB}{255, 150, 255}
\definecolor{sidewalkcolorkt}{RGB}{75, 0, 75}
\definecolor{other-groundcolor}{RGB}{75, 0, 175}
\definecolor{buildingcolor}{RGB}{0, 200, 255}
\definecolor{fencecolor}{RGB}{50, 120, 255}
\definecolor{vegetationcolorkt}{RGB}{0, 175, 0}
\definecolor{trunkcolor}{RGB}{0, 60, 135}
\definecolor{terraincolorkt}{RGB}{80, 240, 150}
\definecolor{polecolor}{RGB}{150, 240, 255}
\definecolor{traffic-signcolor}{RGB}{0, 0, 255}

\newcommand{\makecolorcell}[2]{%
  \makecell[b]{%
    \begin{turn}{90}%
      \textcolor{black}{%
        \colorbox{#1}{\hspace{3pt}\rule{0pt}{3pt}}\ #2%
      }%
    \end{turn}%
  }%
}
\newcommand{\best}[1]{\textcolor{black}{\textbf{#1}}}
\newcommand{\second}[1]{\textcolor{black}{\underline{#1}}}

\title{SpikeOcc: Towards Resilient Multimodal 3D Occupancy Perception via Spiking Experts}
\author{Yiding Sun$^{1,2}$,
Zijie Xu$^{1}$,
Hongqiang Lin$^{3}$,
Jiajun Cao$^{1}$,
Dongxu Zhang$^{4,2}$,
Kai Tang$^{1}$,
\\\bfseries 
Jihua Zhu$^{2}$,
Wenxuan Liu$^{1}$,
Zhaofei Yu$^{1}$,
Tiejun Huang$^{1}$
\\\rmfamily 
$^{1}$Peking University \quad $^{2}$Xi'an Jiaotong University
\quad $^{3}$Zhejiang University \quad $^{4}$Tsinghua University
}

\iclrfinalcopy
\iclrfinalcopy 
\begin{document}

\maketitle

\begin{abstract}
Multimodal 3D occupancy prediction underpins safe autonomous driving through comprehensive scene understanding. However, existing methods have largely pursued higher accuracy through sophisticated fusion, leaving resilience to missing sensors underexplored. In this paper, we investigate spiking neural networks (SNNs) as a means of building more resilient multimodal perception. Still, the discrete nature of spiking neurons makes optimization challenging. To overcome this difficulty, SpikeOcc is introduced with two complementary components, a Spiking Mixture-of-Experts (SMoE) and Sensor Availability Learning (SAL). Specifically, SMoE organizes RGB, LiDAR, and joint fusion into specialized pathways, enabling full-width unimodal representations while retaining the original multimodal features via a bounded residual. Beyond sensor dropout, SAL employs a staged curriculum with explicit routing supervision, progressively improving robustness under severe sensing conditions. Remarkably, SAL also yields notable gains for conventional models, highlighting its broader potential for resilient multimodal learning. Extensive experiments demonstrate that SpikeOcc maintains competitive full-sensor accuracy while markedly raising the performance floor under sensor outages. SpikeOcc illustrates the promise of neuromorphic computing, pointing toward more reliable and energy-efficient autonomous driving.

\end{abstract}

\section{Introduction}
\vspace{-6pt}
\begin{figure*}[h]
\centering
\includegraphics[width=\linewidth]{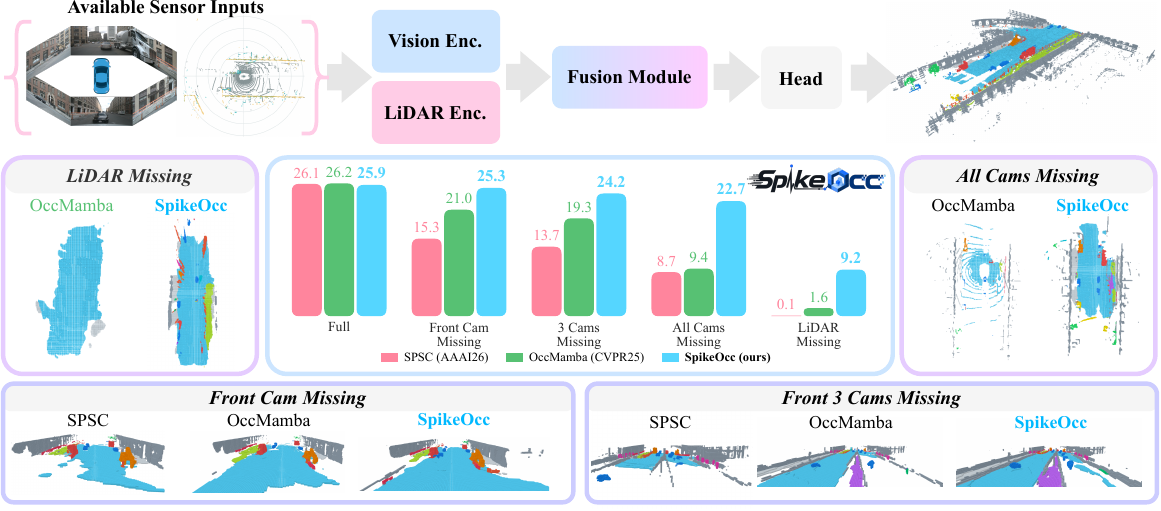} 
\caption{\textbf{Full-sensor accuracy does not guarantee resilience to sensor failures.} Existing MMOMs degrade sharply under modality outages, whereas SpikeOcc remains more resilient.}
\label{fig1}
\end{figure*}
The landscape of 3D occupancy prediction has undergone a paradigm shift from single-sensor perception to tightly coupled multi-sensor fusion~\citep{tian2023occ3d,wang2024occgen,openoccupancy}. Existing multimodal occupancy models (MMOMs) largely share a common pipeline, with major architectural advances concentrated in the fusion module. Yet, behind these advances lies an implicit but stringent assumption: All sensing modalities remain fully available at inference time.

In practice, real-world sensing rarely guarantees such a fixed and complete configuration. Cameras may become unavailable because of communication dropouts or hardware malfunctions, and LiDAR faces the same risks. Recognizing this issue, BEV-based models, such as MetaBEV~\citep{metabev}, have attracted increasing attention. Subsequent studies extend this awareness through unified fusion \citep{unibev}, expert routing~\citep{mome}, and masked-view training \citep{mbev}. 

However, due to its complex scene structures, sensor-level failures remain largely underexplored in 3D occupancy. This leaves a critical question unanswered: \textbf{how does multimodal occupancy behave when part of its sensing stack becomes unavailable?} Fig.~\ref{fig1} reveals a clear failure pattern. A strong MMOM already suffers when the front camera is removed. Once an entire modality disappears, the degradation becomes catastrophic. We refer to this disproportionate collapse as modality outage fragility, raising concerns about the reliability of on-board perception.

\begin{figure*}[t]
\centering
\includegraphics[width=\linewidth]{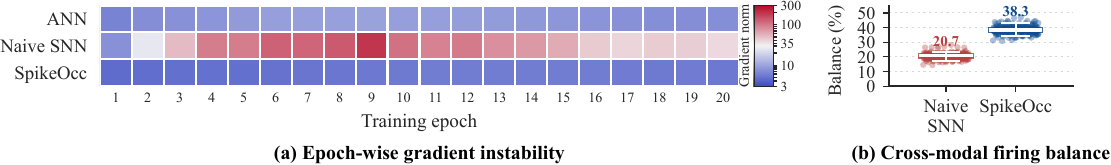} 
\caption{\textbf{Naive spiking fusion exhibits instability.} Firing balance is the lower-rate modality's normalized share and reaches 50\% for equal rates.}
\label{fig2}
\end{figure*}

Spiking neural networks (SNNs) have recently emerged for autonomous driving~\citep{zhu2024sad}, with encouraging robustness against corrupted observations~\citep{kundu2021hiresnn,ding2024robust}. These works motivate us to investigate whether spiking representations can also benefit structured 3D occupancy under changing sensor availability \textit{(See Appendix~\ref{appa})}. Our initial results suggest that the answer is far from straightforward. In Fig.~\ref{fig2}(a), directly converting a strong MMOM into its spiking counterpart causes severe gradient explosion during training. Camera and LiDAR features follow different representation statistics, and spiking conversion further translates this discrepancy into heterogeneous firing dynamics \textit{(See Appendix~\ref{appb})}. Worse still, routing both modalities through a shared fusion pathway entangles these dynamics and makes optimization harder in Fig.~\ref{fig2}(b).

Motivated by these observations, we introduce SpikeOcc, the first spike-driven MMOM for 3D occupancy prediction. SpikeOcc combines two complementary designs. First, we develop a Spiking Mixture-of-Experts (SMoE) that decomposes multimodal fusion into RGB, LiDAR, and joint experts. Each modality is processed along a dedicated spiking pathway, while the joint expert retains cross-modal interaction. This specialization improves naive spiking fusion and preserves fallback capability under sensor outages. Second, we introduce Sensor Availability Learning (SAL), a curriculum that progressively teaches the model to handle diverse sensor states instead of a fixed configuration. Interestingly, equipping conventional MMOMs with SAL further unlocks their resilience, delivering a second leap in performance. Our contributions are fourfold.

\begin{itemize}
    \item We uncover modality outage fragility in MMOMs, revealing a previously overlooked gap between full-sensor accuracy and resilience under sensor failures.
    \item We introduce SpikeOcc, the first spike-driven MMOM for 3D occupancy prediction. Our study establishes spiking computation as a viable representation paradigm and identifies the key difficulty in transferring conventional fusion into the spiking regime. 
    \item We develop SMoE and SAL from architectural and learning perspectives. SMoE specializes heterogeneous representations through spiking experts, while SAL explicitly learns across diverse sensor-availability states.
    \item Extensive experiments show that SpikeOcc performs on par with counterparts under complete sensing, while effectively raising the performance floor under modality outages.
\end{itemize}

\section{Related Work}
\textbf{3D Semantic Occupancy Prediction.} Early efforts, such as LMSCNet~\citep{roldao2020lmscnet} and 
JS3C-Net~\citep{yan2021sparse}, were built primarily on LiDAR, whose sparse 
observations and sensing cost motivated a shift toward vision-centric occupancy 
perception. MonoScene~\citep{cao2022monoscene} and 
SurroundOcc~\citep{wei2023surroundocc} established representative monocular 
and surround-view formulations, followed by stronger architectures such as 
TPVFormer~\citep{huang2023tpvformer}, OccFormer~\citep{zhang2023occformer}, 
and VoxFormer~\citep{li2023voxformer}, richer rendering strategies such as 
UniOcc~\citep{pan2023uniocc} and RenderOcc~\citep{pan2024renderocc}, and 
more compact representations such as COTR~\citep{ma2024cotr}, 
STCOcc~\citep{liao2025stcocc}, and Dr.Occ~\citep{zhu2026drocc}. 
More recently, resilience to sensor failures has emerged as an important concern 
in deployable perception~\citep{orr2021radar,almalioglu2022robust}. UniBEV~\citep{unibev}, MetaBEV~\citep{metabev}, and 
M-BEV~\citep{mbev} target missing or degraded 
observations in BEV-oriented perception, while FlexMap~\citep{wang2026flexmap}, Uni-Map~\citep{hao2026unimap}, 
and SafeMap~\citep{hao2025safemap} extend related ideas to map construction. M2-Occ~\citep{lin2026m2occ} specifically investigates 3D semantic occupancy prediction under incomplete multi-camera observations. Despite this progress, resilience across heterogeneous sensor availability 
remains underexplored in dense semantic occupancy. In this work, we take a pioneering step in utilizing SNNs for resilient occupancy learning, aiming to 
bridge this gap and advance multimodal occupancy inference.

\textbf{Spiking Neural Networks.} As the third generation of neural networks, spiking neural networks (SNNs) 
have shown remarkable progress across a wide range of tasks, achieving 
performance comparable to that of ANNs~\citep{hu2023spiking}. 
Their initial applications span image classification~\citep{yao2023spikedriven,yao2025scaling}, 
object detection~\citep{luo2024integer,li2025brain,qu2025spiking},
semantic segmentation~\citep{lei2025spike2former}, 
video classification~\citep{zou2025spikevideoformer}, 
reinforcement learning~\citep{xu2025proxy,xu2026carebn}, 
temporal prediction~\citep{dong2026prednext}, 
object tracking~\citep{zhang2026spiketrack,shan2026sdtrack,sun2026spikingmot}, 
and event-driven recognition~\citep{liu2021event,liu2025sota}. 
In parallel, SNNs have officially entered the era of 3D vision~\citep{sun2025hyperpoint,sun2026curve3d,sun2026tri,sun2026align}, 
with representative works including SpikePEFT~\citep{guo2026spikepeft}, 
Spiking Discrepancy Transformer~\citep{lu2026spiking}, 
SPM~\citep{wu2025efficient}, and 3DSMT~\citep{zhou2026smt}. 
However, so far, there has been no exciting baseline in the 3D occupancy domain. 
This study is dedicated to investigating the potential contributions of SNNs 
to this task, further advancing neuromorphic visual perception.

\section{Method}
\subsection{Problem Formulation}

We consider multimodal 3D semantic occupancy prediction under incomplete
sensing. Given $N_c$ synchronized camera images
$\mathcal{I}=\{I^n\}_{n=1}^{N_c}$ and an aggregated LiDAR point cloud
$\mathcal{P}$, the goal is to predict the occupancy state and semantic label
of each voxel in a predefined 3D space $\Omega$. Let
$\mathcal{Y}=\{0,\ldots,K-1\}$ denote the label set, where $0$ represents
free space. The prediction is
$\hat{\mathbf{Y}}\in\mathcal{Y}^{|\Omega|}$. To describe sensor availability, we introduce a binary view mask
$\mathbf{m}\in\{0,1\}^{N_c}$ and a LiDAR availability indicator
$m_l\in\{0,1\}$. The indicator and ratio are
defined as
\begin{equation}
m_c=\mathbb{I}\!\left[\sum_{n=1}^{N_c}m_n>0\right],
\qquad
r_c=\frac{1}{N_c}\sum_{n=1}^{N_c}m_n.
\end{equation}  
The resulting availability descriptor is
$\mathbf{a}=(\mathbf{m},m_c,m_l,r_c)$. A sensor configuration
$s\in\mathcal{S}$ determines a masking transformation $T_s$, yielding
the available observations $\mathcal{X}_s=T_s(\mathcal{I},\mathcal{P})$.
We require $m_c+m_l\geq1$, so at least one modality remains available.

The objective is to learn a single predictor
\begin{equation}
f_\Theta:(\mathcal{X}_s,\mathbf{a}_s)
\longmapsto \hat{\mathbf{Y}}_s,
\qquad s\in\mathcal{S},
\end{equation}
that maintains accurate occupancy prediction across different sensing
conditions without retraining a separate model for each configuration.
Unlike conventional training~\citep{occmamba}, our learning objective must account for the risks associated with different sensing conditions.
\begin{equation}
R_s(\Theta)=
\mathbb{E}_{(\mathcal{X},\mathbf{Y})\sim\mathcal{D}}
\left[
\mathcal{L}_{\mathrm{occ}}
\big(f_\Theta(T_s(\mathcal{X}),\mathbf{a}_s),\mathbf{Y}\big)
\right].
\end{equation}
Sensor Availability Learning (Sec.~\ref{sal}) will later control how these risks are sampled during training.

\begin{figure*}[t]
\centering
\includegraphics[width=\linewidth]{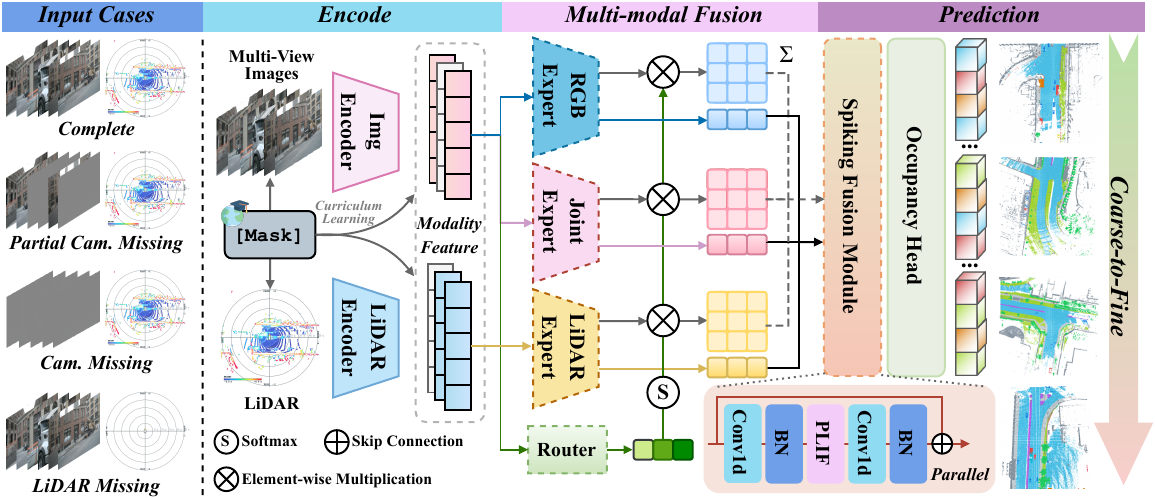} 
\caption{\textbf{Overview of SpikeOcc.} Masked camera and LiDAR inputs are encoded and processed by specialized experts. Router-weighted features pass through the spiking fusion module and occupancy head, while SAL progressively introduces missing-sensor cases during training.
}
\label{fig3}
\end{figure*}

\subsection{Spiking Mixture-of-Experts (SMoE)}

\textbf{Width-first Decomposition.}
Fig.~\ref{fig3} illustrates the proposed SpikeOcc architecture.
The camera and LiDAR encoders first produce spatially aligned voxel
features $\mathbf{C}_0$ and $\mathbf{L}_0$. After availability masking
and projection, we obtain
\begin{equation}
\begin{aligned}
\widetilde{\mathbf{C}}_0 &= m_c\mathbf{C}_0,
&
\mathbf{C} &= \phi_c(\widetilde{\mathbf{C}}_0),\\
\widetilde{\mathbf{L}}_0 &= m_l\mathbf{L}_0,
&
\mathbf{L} &= \phi_l(\widetilde{\mathbf{L}}_0),
\end{aligned}
\end{equation}
where $\phi_c$ and $\phi_l$ are independent $1\times1\times1$
convolutional projections. Both projected features have width $d$,
and the fusion interface has width $D=2d$.

Rather than simply deepening a shared fusion network~\citep{Nspikemoe}, SMoE develops
three parallel experts: an RGB expert $E_c$, a LiDAR expert $E_l$,
and a joint expert $E_j$. Each produces a $D$-channel representation.
This design addresses two potential failure modes. Half-width
unimodal outputs would reduce the input capacity of the occupancy
predictor when one modality disappears~\citep{zagoruyko2016wide}, whereas an unconstrained
joint transformation could overwrite the original modality features~\citep{wang2023shaspec}.
SMoE therefore equips unimodal experts with full-width outputs and
retains the original feature concatenation as continuous foundation.

\textbf{Spiking Residual Generation.}
Each expert contains an independent spiking bottleneck with the same
operator structure. We use Parametric Leaky Integrate-and-Fire
(PLIF) neurons~\citep{fang2021plif}. For an input current
$\mathbf{x}^{t}$, the membrane dynamics are
\begin{equation}
\begin{aligned}
\widetilde{\mathbf{u}}^{t}
&=(1-\beta)\mathbf{u}^{t-1}+\beta\mathbf{x}^{t},
&
\beta&=\sigma(\theta_\beta),\\
\mathbf{s}^{t}
&=\mathbb{H}(\widetilde{\mathbf{u}}^{t}-V_{\mathrm{th}}),
&
\mathbf{u}^{t}
&=\widetilde{\mathbf{u}}^{t}-V_{\mathrm{th}}\mathbf{s}^{t},
\end{aligned}
\label{eq5}
\end{equation}
where $\theta_\beta$ is learnable, $\mathbb{H}$ is the hard-threshold
function, and the last line implements soft reset~\citep{rueckauer2017conversion}.  During backpropagation, the non-differentiable
threshold is approximated by the sigmoid surrogate~\citep{wu2018spatio,neftci2019surrogate}
\begin{equation}
\frac{\partial \mathbf{s}^{t}}
{\partial\widetilde{\mathbf{u}}^{t}}
\approx
\alpha\sigma(\alpha z)
\bigl(1-\sigma(\alpha z)\bigr),
\qquad
z=\widetilde{\mathbf{u}}^{t}-V_{\mathrm{th}}.
\end{equation}
Here, we set $\alpha=4$ throughout all experiments. The spiking bottleneck compresses the input to $D/4$ channels, extracts local 3D context, and expands the result back to $D$:
\begin{equation}
\begin{aligned}
\mathbf{s}_1^t
&=\operatorname{PLIF}_1(\mathcal{P}_1(\mathbf{X})),
&
\mathbf{s}_2^t
&=\operatorname{PLIF}_2(\mathcal{P}_2(\mathbf{s}_1^t)),
&
\mathbf{R}^t
&=\mathcal{P}_3(\mathbf{s}_2^t),
&
\overline{\mathbf{R}}
&=\frac{1}{T}\sum_{t=1}^{T}\mathbf{R}^t.
\end{aligned}
\end{equation}
Here, $\mathcal{P}_1$ and $\mathcal{P}_3$ are convolutional projections, and $\mathcal{P}_2$ is a depthwise projection, all followed by batch normalization. The PLIF layers generate binary spikes, and the final projection produces a continuous readout. We average $T$ internal steps and reset membrane states between samples.

\textbf{Spiking Experts.}
For the two unimodal experts, continuous projections preserve the original information while spiking branches learn residual
corrections:
\begin{equation}
\mathbf{F}_c
=\psi_c(\mathbf{C})+\gamma_c\overline{\mathbf{R}}_c,
\qquad
\mathbf{F}_l
=\psi_l(\mathbf{L})+\gamma_l\overline{\mathbf{R}}_l.
\end{equation}
Here $\psi_c,\psi_l$ project the modality features to width $D$,
and $\gamma_c,\gamma_l$ are learnable scalars initialized to $0.01$.
Thus, both experts initially remain close to stable continuous
projections while learning spiking corrections. The joint expert instead preserves the original masked
concatenation
$\mathbf{B}=[\widetilde{\mathbf{C}}_0;
\widetilde{\mathbf{L}}_0]$ and learns a bounded residual:
\begin{equation}
\mathbf{F}_j
=\mathbf{B}+\rho\overline{\mathbf{R}}_j,
\qquad
\rho=\rho_{\max}\sigma(\theta_\rho).
\end{equation}
The joint spiking branch receives $[\mathbf{C};\mathbf{L}]$.
We set $\rho_{\max}=0.2$ and initialize $\rho=0.01$.
The bounded coefficient limits the residual gain, reducing
the risk of large perturbations during early optimization.

\textbf{Routing.}
The router combines global feature statistics with explicit
availability information. Let
$\mathbf{g}_c=\operatorname{GAP}(\mathbf{C})$ and
$\mathbf{g}_l=\operatorname{GAP}(\mathbf{L})$. $\operatorname{GAP}(\cdot)$ denotes global average pooling over the spatial
dimensions. A two-layer MLP predicts
three logits from
\begin{equation}
\mathbf{h}=
[\mathbf{g}_c;\mathbf{g}_l;m_c;m_l;r_c],
\qquad
\mathbf{z}=\operatorname{MLP}(\mathbf{h})\in\mathbb{R}^{3}.
\end{equation}
The availability vector
$\mathbf{q}=(m_c,m_l,m_cm_l)$ excludes experts that require an
unavailable modality. The remaining logits are normalized as
\begin{equation}
w_k =
\frac{q_k \exp(z_k)}
{\sum_{i\in\{c,l,j\}} q_i\exp(z_i)},
\qquad
\mathbf{F} =
\sum_{k\in\{c,l,j\}} w_k\mathbf{F}_k.
\label{eq:router_weights}
\end{equation}
\subsection{Sensor Availability Learning}
\label{sal}
\begin{wrapfigure}[10]{r}{0.63\columnwidth}
    \vspace{-12pt}
    \centering
    \begin{minipage}[t]{0.495\linewidth}
        \vspace{0pt}
        \centering
        \includegraphics[
            width=\linewidth,
            height=0.16\textheight,
            keepaspectratio
        ]{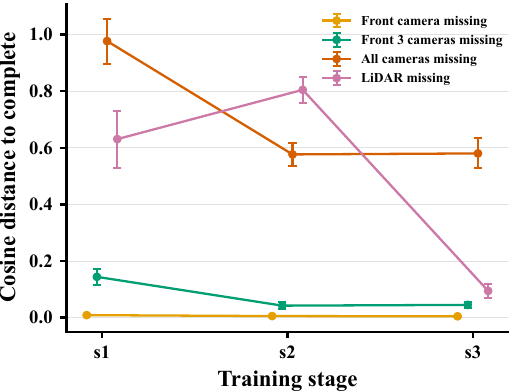}

        \captionof{figure}{Representation adaptation across training stages.}
        \label{fig:sal_representation}
    \end{minipage}%
    \hfill
    \begin{minipage}[t]{0.495\linewidth}
        \vspace{0pt}
        \centering
        \includegraphics[
            width=\linewidth,
            height=0.16\textheight,
            keepaspectratio
        ]{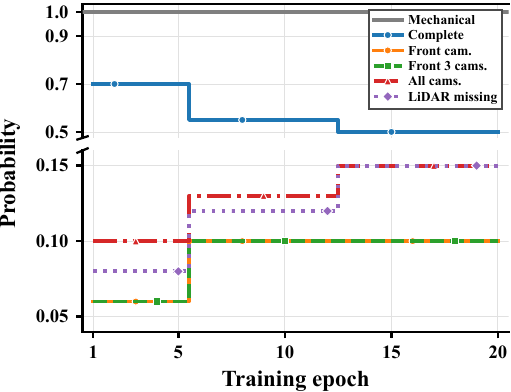}

        \captionof{figure}{Curriculum learning schedule in SAL.}
        \label{fig:sal_curriculum}
    \end{minipage}

    \vspace{-8pt}
\end{wrapfigure}

\textbf{Mask Propagation.}
SAL trains the same model across a structured set of sensing states. A sampled state is propagated consistently through the network.
For the camera branch, the view mask is applied to the lifted
frustum features before voxel pooling:
\begin{equation}
\begin{aligned}
\widetilde{\mathbf{V}}_{n,u,k}
&=m_n\,p(d_k\mid u,n)\mathbf{g}_{u,n},\\
\mathbf{C}_0(v)
&=\sum_{\pi(n,u,d_k)=v}
\widetilde{\mathbf{V}}_{n,u,k}.
\end{aligned}
\end{equation}
Here $p(d_k\mid u,n)$ is the predicted depth distribution,
$\mathbf{g}_{u,n}$ is the image context feature, and $\pi$ maps
a frustum sample to its 3D voxel. The same view mask excludes
unavailable cameras from depth supervision,
ensuring feature construction describes the same sensing state~\citep{philion2020lift}.

\textbf{Curriculum Learning.}
Fig.~\ref{fig:sal_representation} reveals that direct training
with sensor dropout can lead to fluctuating representation distances between complete and incomplete inputs.
To stabilize adaptation, we introduce a
three-stage curriculum that gradually increases the exposure to
sensor failures.
For each sample, the sensor state is drawn from a distribution that
varies with the training epoch:
\begin{equation}
s_b\sim\operatorname{Categorical}(\mathbf{p}_e),
\qquad
\mathbf{p}_e=\mathbf{p}^{(k)},\quad
e\in\mathcal{E}_k,\quad k\in\{1,2,3\}.
\end{equation}
Here, $e$ denotes the current epoch, $\mathcal{E}_k$ the epoch range of stage $k$, and $\mathbf{p}^{(k)}$ the corresponding sampling distribution. Fig.~\ref{fig:sal_curriculum} illustrates the resulting curriculum.
The curriculum progresses from complete sensing to failures. Alternative orderings are analyzed in Appendix~\ref{appc}.

\textbf{Learning Objective.}
To encourage the intended expert roles, we introduce a lightweight routing objective. Let $y^{\mathrm{route}}$ denote the target defined in \textit{Appendix~\ref{appd}}. We supervise
the raw router logits:
\begin{equation}
\mathcal{L}_{\mathrm{route}}
=\lambda_r\operatorname{CE}
(\mathbf{z},y^{\mathrm{route}}),
\qquad \lambda_r=0.05.
\label{eq:route_loss}
\end{equation}
This auxiliary loss therefore teaches the router the responsibilities without replacing the soft mixture
among valid experts. The occupancy objective remains unchanged~\citep{occmamba}. At each coarse and fine
scale, it combines class-weighted cross-entropy, semantic scaling,
geometric scaling~\citep{cao2022monoscene}, and Lovasz-Softmax losses~\citep{berman2018lovasz}:
\begin{equation}
\mathcal{L}_{\mathrm{occ}}
=
\lambda_1\mathcal{L}_{\mathrm{CE}}
+\lambda_2\mathcal{L}_{\mathrm{sem}}
+\lambda_3\mathcal{L}_{\mathrm{geo}}
+\lambda_4\mathcal{L}_{\mathrm{Lovasz}}.
\end{equation}
The inherited depth loss~\citep{li2023bevdepth} is evaluated only
on valid camera views. The total training loss is
\begin{equation}
\mathcal{L}_{\mathrm{total}}
=
\mathcal{L}_{\mathrm{occ}}
+\mathcal{L}_{\mathrm{depth}}
+\mathcal{L}_{\mathrm{route}}.
\end{equation}
Here, $\lambda_1$--$\lambda_4$ balance the contributions of the
occupancy loss terms and are all empirically set to $1$.
\section{Experiments}
\begin{table*}[t]
\centering
\caption{Quantitative comparisons on the OpenOccupancy validation set. C, D, and L denote camera, depth, and LiDAR, respectively. Within each sensor setting, the \textbf{best}, \underline{second-best}, and \textcolor{tablegray}{remaining results} are shown in bold, underlined, and gray, respectively. For fair comparison, results marked with $\dagger$ are reproduced by us and averaged over 3 seeds. Same for subsequent tables.}
\resizebox{\linewidth}{!}{
\setlength{\tabcolsep}{1mm}
\begin{threeparttable}
\begin{tabular}{l|c|
>{\color{tablegray}}c
>{\color{tablegray}}c|
*{16}{>{\color{tablegray}}c}}
\toprule

\makecell[b]{Method} &
\makecell[b]{Input\\Modality} &
\textcolor{black}{\makecell[b]{IoU}} &
\textcolor{black}{\makecell[b]{mIoU}} &
\makecolorcell{barriercolor}{barrier} &
\makecolorcell{bicyclecolor}{bicycle} &
\makecolorcell{buscolor}{bus} &
\makecolorcell{carcolor}{car} &
\makecolorcell{construction_vehiclecolor}{const. veh.} &
\makecolorcell{motorcyclecolor}{motorcycle} &
\makecolorcell{pedestriancolor}{pedestrian} &
\makecolorcell{traffic_conecolor}{traffic cone} &
\makecolorcell{trailercolor}{trailer} &
\makecolorcell{truckcolor}{truck} &
\makecolorcell{driveable_surfacecolor}{drive surf.} &
\makecolorcell{other_flatcolor}{other\_flat} &
\makecolorcell{sidewalkcolor}{sidewalk} &
\makecolorcell{terraincolor}{terrain} &
\makecolorcell{manmadecolor}{manmade} &
\makecolorcell{vegetationcolor}{vegetation} \\
\midrule

\multicolumn{20}{@{}c@{}}{
\cellcolor{settingfull}
\hspace{3pt}\textit{All Sensors Available}\hspace{3pt}
}\\[-1pt]

MonoScene~\citep{cao2022monoscene} & C
& 18.4 & 6.9
& 7.1 & 3.9 & 9.3 & 7.2
& 5.6 & 3.0 & 5.9 & 4.4
& 4.9 & 4.2 & 14.9 & 6.3
& 7.9 & 7.4 & 10.0 & 7.6 \\

JS3C-Net~\citep{yan2021sparse} & L
& 30.2 & 12.5
& 14.2 & 3.4 & 13.6 & 12.0
& 7.2 & 4.3 & 7.3 & 6.8
& 9.2 & 9.1 & 27.9 & 15.3
& 14.9 & 16.2 & 14.0 & 24.9 \\

3DSketch~\citep{3dsketch} & C\&D
& 25.6 & 10.7
& 12.0 & 5.1 & 10.7 & 12.4
& 6.5 & 4.0 & 5.0 & 6.3
& 8.0 & 7.2 & 21.8 & 14.8
& 13.0 & 11.8 & 12.0 & 21.2 \\

M-CONet~\citep{openoccupancy} & C\&L
& 29.5 & 20.1
& 23.3 & 13.3 & 21.2 & 24.3
& 15.3 & 15.9 & 18.0 & 13.3
& 15.3 & 20.7 & 33.2 & 21.0
& 22.5 & 21.5 & 19.6 & 23.2 \\

Co-Occ~\citep{pan2024coocc} & C\&L
& 30.6 & 21.9
& 26.5 & 16.8 & 22.3 & 27.0
& 10.1 & 20.9 & 20.7 & 14.5
& 16.4 & 21.6 & 36.9 & 23.5
& 5.5 & 23.7 & 20.5 & 23.5 \\

OccMamba~\citep{occmamba} & C\&L
& \second{35.7} & \best{26.2}
& 30.2 & \second{20.5} & \best{26.5} & 29.5
& \best{18.8} & \best{26.0} & 23.7
& \second{19.9} & \best{20.6} & \best{25.4}
& \second{38.4} & 26.5
& \second{27.0} & \second{26.6}
& \best{28.9} & \second{30.5} \\

RIOcc~\citep{fan2025riocc} & C\&L
& 35.4 & 25.9
& 30.2 & 19.8 & \second{25.8} & 28.7
& \second{18.3} & 24.8
& \best{31.8} & \best{21.8}
& \second{20.5} & 24.9
& 37.2 & 24.5
& 25.5 & 24.9
& 27.0 & 28.8 \\

SPSC~\citep{spsc} & C\&L
& 35.2 & \second{26.1}
& \best{31.4} & \best{20.6}
& 25.1 & \best{30.6}
& 15.7 & \second{25.9}
& \second{27.1} & 18.8
& 19.7 & \second{25.2}
& \second{38.4} & \second{26.9}
& 26.8 & 26.5
& 28.5 & 30.4 \\

\textbf{SpikeOcc (ours)} & C\&L
& \best{35.8} & 25.9
& \second{30.8} & 19.4
& 25.4 & \second{30.2}
& 13.7 & 25.3
& 26.5 & 17.7
& 19.5 & 25.0
& \best{39.9} & \best{27.1}
& \best{27.5} & \best{27.0}
& \second{28.6} & \best{30.8} \\

\midrule[0.7pt]

\multicolumn{20}{@{}c@{}}{
\cellcolor{settingfront}
\hspace{3pt}\textit{Front Camera Missing}\hspace{3pt}
}\\[-1pt]

OccMamba$^\dagger$~\citep{occmamba} & C\&L
& \second{30.9} & \second{21.0}
& \second{26.5} & \second{11.5}
& \second{20.6} & \second{26.0}
& \second{12.1} & \second{18.8}
& \second{23.7} & \second{12.5}
& \second{15.0} & \second{19.5}
& \second{35.2} & \second{20.6}
& \second{23.7} & \second{22.9}
& \second{24.1} & 23.3 \\

SPSC$^\dagger$~\citep{spsc} & C\&L
& 22.3 & 15.3
& 15.5 & 9.0
& 16.1 & 22.2
& 8.0 & 10.7
& 20.8 & 12.0
& 11.7 & 17.9
& 16.4 & 10.1
& 12.0 & 13.3
& \second{24.1} & \second{25.2} \\

\textbf{SpikeOcc (ours)} & C\&L
& \best{36.2} & \best{25.3}
& \best{30.1} & \best{17.7}
& \best{24.2} & \best{29.5}
& \best{15.1} & \best{21.5}
& \best{25.9} & \best{17.6}
& \best{19.6} & \best{24.5}
& \best{39.7} & \best{25.9}
& \best{27.0} & \best{26.5}
& \best{28.5} & \best{30.9} \\

\midrule[0.7pt]

\multicolumn{20}{@{}c@{}}{
\cellcolor{settingfrontthree}
\hspace{3pt}\textit{Front-3 Cameras Missing}\hspace{3pt}
}\\[-1pt]

OccMamba$^\dagger$~\citep{occmamba} & C\&L
& \second{30.4} & \second{19.3}
& \second{25.1} & \second{8.8}
& \second{19.1} & \second{23.8}
& \second{10.0} & \second{14.6}
& \second{20.9} & \second{10.6}
& \second{13.7} & \second{16.8}
& \second{34.8} & \second{19.9}
& \second{23.0} & \second{21.9}
& 23.7 & 22.6 \\

SPSC$^\dagger$~\citep{spsc} & C\&L
& 21.9 & 13.7
& 13.6 & 6.2
& 13.4 & 18.9
& 6.0 & 8.2
& 18.7 & 9.9
& 10.0 & 15.6
& 16.2 & 9.6
& 11.6 & 12.3
& \second{23.8} & \second{24.9} \\

\textbf{SpikeOcc (ours)} & C\&L
& \best{36.3} & \best{24.2}
& \best{29.0} & \best{14.3}
& \best{23.1} & \best{28.9}
& \best{14.3} & \best{18.6}
& \best{24.4} & \best{15.9}
& \best{18.5} & \best{23.3}
& \best{39.7} & \best{25.3}
& \best{26.6} & \best{25.9}
& \best{28.4} & \best{30.8} \\

\midrule[0.7pt]

\multicolumn{20}{@{}c@{}}{
\cellcolor{settingallcam}
\hspace{3pt}\textit{All Cameras Missing}\hspace{3pt}
}\\[-1pt]

OccMamba$^\dagger$~\citep{occmamba} & C\&L
& 18.8 & \second{9.4}
& 3.9 & \second{1.3}
& \second{10.0} & \second{9.4}
& \second{6.4} & \second{3.3}
& \second{16.8} & \second{3.3}
& \second{6.3} & 5.6
& \second{17.7} & \second{10.0}
& \second{10.5} & 8.6
& 19.6 & 18.2 \\

SPSC$^\dagger$~\citep{spsc} & C\&L
& \second{21.8} & 8.7
& \second{6.9} & 1.2
& 4.2 & 5.5
& 0.4 & 2.8
& 13.8 & 0.9
& 4.6 & \second{8.7}
& 15.6 & 7.0
& 10.4 & \second{9.4}
& \second{23.3} & \second{24.2} \\

\textbf{SpikeOcc (ours)} & C\&L
& \best{36.8} & \best{22.7}
& \best{27.3} & \best{10.1}
& \best{21.2} & \best{27.6}
& \best{12.1} & \best{15.4}
& \best{22.0} & \best{13.2}
& \best{17.4} & \best{21.1}
& \best{40.3} & \best{24.0}
& \best{26.1} & \best{25.4}
& \best{28.5} & \best{31.1} \\

\midrule[0.7pt]

\multicolumn{20}{@{}c@{}}{
\cellcolor{settinglidar}
\hspace{3pt}\textit{LiDAR Missing}\hspace{3pt}
}\\[-1pt]

OccMamba$^\dagger$~\citep{occmamba} & C\&L
& \second{7.8} & \second{1.6}
& 0.0 & 0.0
& \second{0.1} & 0.0
& 0.0 & 0.0
& 0.0 & 0.0
& 0.0 & 0.0
& \second{14.5} & \second{3.2}
& \second{4.7} & \second{3.6}
& 0.0 & 0.0 \\

SPSC$^\dagger$~\citep{spsc} & C\&L
& 0.2 & 0.1
& 0.0 & 0.0
& 0.0 & 0.0
& 0.0 & 0.0
& 0.0 & 0.0
& 0.0 & 0.0
& 0.8 & 0.4
& 0.4 & 0.0
& 0.0 & 0.0 \\

\textbf{SpikeOcc (ours)} & C\&L
& \best{19.7} & \best{9.2}
& \best{7.9} & \best{4.5}
& \best{8.4} & \best{12.7}
& \best{2.2} & \best{6.3}
& \best{7.7} & \best{5.1}
& \best{0.8} & \best{7.9}
& \best{30.6} & \best{15.8}
& \best{16.1} & \best{14.6}
& \best{1.9} & \best{5.4} \\

\bottomrule
\end{tabular}
\end{threeparttable}
}
\label{table_openoccupancy}
\end{table*}

\subsection{Dataset and Metrics}
\textbf{Datasets.}
We evaluate SpikeOcc on OpenOccupancy~\citep{openoccupancy} 
and SemanticKITTI~\citep{behley2019semantickitti}. 
OpenOccupancy extends nuScenes with dense semantic occupancy annotations, 
containing 28,130 training frames and 6,019 validation frames with 17 semantic 
categories. Its evaluation space covers $[-51.2, 51.2]$ m along the $X$ and $Y$ 
axes and $[-3, 5]$ m along the $Z$ axis, with a voxel resolution of $0.2$\,m.
SemanticKITTI provides large-scale LiDAR sequences for semantic scene completion 
with 19 semantic categories. Following the official protocol, sequences 
00--07 and 09--10 are used for training, sequence 08 for validation, and 
sequences 11--21 for testing. The scene is voxelized into a 
$256\times256\times32$ grid at a resolution of $0.2$\,m.

\textbf{Metrics.}
We report Intersection over 
Union (IoU) for geometric occupancy and mean Intersection over Union (mIoU) 
for semantic occupancy. IoU measures occupied-versus-empty prediction quality, 
while mIoU averages the class-wise IoU over all semantic categories~\citep{song2017semantic}. Additional metric definitions are provided in \textit{Appendix~\ref{appe}}.

\subsection{Implementation Details}
\textbf{Architecture and Training.} Our implementation is built upon MMDetection3D~\citep{mmdet3d2020}. 
For the camera branch, we adopt an ImageNet-pretrained ResNet-50~\citep{he2016deep} 
as the image backbone and resize the input images to $256\times704$. 
For the LiDAR branch, we aggregate 10 LiDAR sweeps and encode the resulting 
point clouds with a voxel-based encoder~\citep{yan2018second}. The feature dimension is set to 384 
throughout our experiments. Our spiking modules use both LIF~\citep{gerstner2002spiking} and PLIF~\citep{fang2021plif} neurons. PLIF is used in default. We initialize
$\tau=2$, equivalently $\beta=0.5$, and use
$V_{\mathrm{th}}=0.5$. We train all models using AdamW with a weight decay of 0.01 and an initial 
learning rate of $3\times10^{-4}$. Training lasts for 20 epochs with a 
1,500-iteration warm-up, and gradient clipping is applied to stabilize 
optimization. All experiments are conducted on eight NVIDIA A800 GPUs with 
a global batch size of 8. The training run takes approximately 46 hours. The inference protocol is detailed in \textit{Appendix~\ref{appf}}. 

\textbf{Sensor Configurations.} To evaluate resilience under varying sensor availability, we consider five 
sensor configurations on OpenOccupancy
and three configurations on SemanticKITTI. Unless otherwise specified, all methods are evaluated under identical sensor configurations, while training follows their respective default protocols.

\begin{table*}[t]
\centering
\caption{Performance on the SemanticKITTI test set. Within each sensor setting, the \textbf{best}, \underline{second-best}, and \textcolor{tablegray}{remaining results} are shown in bold, underlined, and gray, respectively.}
\label{table_semantikkiti_full}

\resizebox{\linewidth}{!}{%
\setlength{\tabcolsep}{0.5mm}
\begin{threeparttable}
\begin{tabular}{l|c|>{\color{tablegray}}c>{\color{tablegray}}c|*{19}{>{\color{tablegray}}c}}
\toprule

\textcolor{black}{\makecell[b]{Method}} &
\textcolor{black}{\makecell[b]{Input\\Modality}} &
\textcolor{black}{\makecell[b]{IoU}} &
\textcolor{black}{\makecell[b]{mIoU}} &
\makecolorcell{roadcolor}{road} &
\makecolorcell{sidewalkcolorkt}{sidewalk} &
\makecolorcell{parkingcolor}{parking} &
\makecolorcell{other-groundcolor}{other ground} &
\makecolorcell{buildingcolor}{building} &
\makecolorcell{carcolorkt}{car} &
\makecolorcell{truckcolorkt}{truck} &
\makecolorcell{bicyclecolorkt}{bicycle} &
\makecolorcell{motorcyclecolorkt}{motorcycle} &
\makecolorcell{other-vehiclecolorkt}{other vehicle} &
\makecolorcell{vegetationcolorkt}{vegetation} &
\makecolorcell{trunkcolor}{trunk} &
\makecolorcell{terraincolorkt}{terrain} &
\makecolorcell{personcolor}{person} &
\makecolorcell{bicyclistcolor}{bicyclist} &
\makecolorcell{motorcyclistcolor}{motorcyclist} &
\makecolorcell{fencecolor}{fence} &
\makecolorcell{polecolor}{pole} &
\makecolorcell{traffic-signcolor}{traffic sign} \\
\midrule

\multicolumn{23}{@{}c@{}}{\cellcolor{settingall}\hspace{3pt}\textit{All Sensors Available}\hspace{3pt}}\\[-1pt]

SurroundOcc~\citep{wei2023surroundocc} & C & 34.7 & 11.9 & 56.9 & 28.3 & \second{30.2} & \second{6.8} & 15.2 & 20.6 & 1.4 & 1.6 & 1.2 & 4.4 & 14.9 & 3.4 & 19.3 & 1.4 & 2.0 & 0.1 & 11.3 & 3.9 & 2.4 \\

OccFormer~\citep{zhang2023occformer} & C & 34.5 & 12.3 & 55.9 & 30.3 & \best{31.5} & 6.5 & 15.7 & 21.6 & 1.2 & 1.5 & 1.7 & 3.2 & 16.8 & 3.9 & 21.3 & 2.2 & 1.1 & \second{0.2} & 11.9 & 3.8 & 3.7 \\

RenderOcc~\citep{pan2024renderocc} & C & N/A & 12.8 & 57.2 & 28.4 & 16.1 & 0.9 & 18.2 & 24.9 & 6.0 & 0.4 & 0.3 & 3.7 & 26.2 & 4.9 & 3.6 & 1.9 & \best{3.1} & 0.0 & 9.1 & 6.2 & 3.4 \\

M-CONet~\citep{openoccupancy} & C\&L & N/A & 20.4 & 60.6 & \best{36.1} & 29.0 & \best{13.0} & \second{38.4} & 33.8 & 4.7 & 3.0 & 2.2 & 5.9 & \second{41.5} & \best{20.5} & 35.1 & 0.8 & 2.3 & \best{0.6} & \best{26.0} & 18.7 & 15.7 \\

OccGen$^\dagger$~\citep{wang2024occgen} & C\&L & \second{59.9} & 21.1 & 63.1 & 33.0 & 17.3 & 0.3 & 38.1 & 44.1 & 17.4 & 3.8 & \second{2.9} & 7.9 & 41.4 & \best{20.5} & 46.6 & \second{4.1} & 2.1 & 0.0 & 13.8 & \second{28.6} & \best{16.2} \\

OccMamba$^\dagger$~\citep{occmamba} & C\&L & 59.6 & \best{21.6} & \best{65.1} & \second{35.1} & 16.5 & 0.9 & 38.3 & \second{44.6} & 21.3 & \best{4.4} & \best{3.5} & 8.8 & 41.4 & 19.0 & 45.6 & \best{4.5} & 2.0 & 0.0 & \second{14.0} & \best{29.3} & \second{16.0} \\

SPSC$^\dagger$~\citep{spsc} & C\&L & 59.4 & 21.3 & \second{63.4} & 34.5 & 14.5 & 0.2 & 38.0 & 43.7 & \second{22.2} & \second{4.2} & 1.6 & \best{13.5} & 40.5 & \second{19.3} & \second{46.9} & 3.4 & 1.7 & 0.0 & 13.1 & 28.3 & 15.7 \\

\textbf{SpikeOcc (ours)} & C\&L & \best{60.3} & \second{21.5} & 63.1 & 34.4 & 14.7 & 1.1 & \best{39.0} & \best{45.6} & \best{24.8} & 2.0 & 2.2 & \second{9.2} & \best{42.0} & 18.8 & \best{47.1} & \second{4.1} & \second{2.6} & 0.0 & 13.4 & \best{29.3} & 14.5 \\

\midrule[0.8pt]

\multicolumn{23}{@{}c@{}}{\cellcolor{settingallcamkt}\hspace{3pt}\textit{All Cameras Missing}\hspace{3pt}}\\[-1pt]

OccGen$^\dagger$~\citep{wang2024occgen} & C\&L & 59.1 & 19.8 & 55.0 & 30.1 & 12.1 & \second{0.8} & 37.0 & 41.4 & \second{18.0} & \best{5.0} & \best{4.2} & 8.3 & \second{41.6} & 18.4 & 43.0 & \best{4.5} & 2.0 & 0.0 & 13.1 & 26.1 & \second{15.5} \\

OccMamba$^\dagger$~\citep{occmamba} & C\&L & \second{59.7} & \second{20.9} & 62.6 & 32.6 & \best{16.4} & 0.3 & \second{37.9} & \second{43.7} & 16.4 & 3.7 & \second{2.8} & 7.5 & 41.3 & \best{20.2} & 46.3 & \second{4.1} & \second{2.1} & 0.0 & \best{13.7} & \second{28.5} & \best{16.0} \\

SPSC$^\dagger$~\citep{spsc} & C\&L & 58.9 & 20.7 & \best{63.9} & \second{34.1} & \second{15.4} & 0.2 & 37.0 & 43.4 & 17.8 & \second{3.8} & 1.5 & \best{11.3} & 39.1 & 18.6 & \best{47.3} & 3.3 & 1.8 & 0.0 & 12.1 & 28.1 & \second{15.5} \\

\textbf{SpikeOcc (ours)} & C\&L & \best{60.4} & \best{21.4} & \second{62.9} & \best{34.4} & 14.7 & \best{1.1} & \best{39.0} & \best{45.6} & \best{24.0} & 2.1 & 2.3 & \second{9.3} & \best{42.2} & \second{18.7} & \second{47.2} & \second{4.1} & \best{2.5} & 0.0 & \second{13.5} & \best{29.3} & 14.5 \\

\midrule[0.8pt]

\multicolumn{23}{@{}c@{}}{\cellcolor{settinglidarkt}\hspace{3pt}\textit{LiDAR Missing}\hspace{3pt}}\\[-1pt]

OccGen$^\dagger$~\citep{wang2024occgen} & C\&L & 4.7 & 0.9 & 8.4 & \second{6.8} & 0.0 & 0.0 & \second{0.4} & 0.0 & 0.0 & 0.0 & 0.0 & 0.0 & 0.6 & 0.0 & 1.7 & 0.0 & 0.0 & 0.0 & 0.0 & 0.0 & 0.0 \\

OccMamba$^\dagger$~\citep{occmamba} & C\&L & 9.3 & \second{1.7} & \second{15.4} & 5.6 & 0.4 & 0.0 & 0.1 & 0.0 & 0.0 & 0.0 & 0.0 & 0.0 & 0.3 & 0.0 & \second{10.0} & 0.0 & 0.0 & 0.0 & 0.0 & 0.0 & 0.0 \\

SPSC$^\dagger$~\citep{spsc} & C\&L & \second{11.1} & 1.5 & 15.0 & 5.2 & \second{1.0} & 0.0 & 0.2 & \second{2.1} & 0.0 & \best{0.1} & 0.0 & \second{0.1} & \second{1.4} & 0.0 & 3.1 & 0.0 & 0.0 & 0.0 & \second{0.2} & 0.0 & 0.0 \\

\textbf{SpikeOcc (ours)} & C\&L & \best{21.8} & \best{4.1} & \best{29.1} & \best{13.8} & \best{5.0} & \best{0.3} & \best{0.7} & \best{7.7} & 0.0 & 0.0 & 0.0 & \best{0.4} & \best{4.3} & \best{0.1} & \best{14.5} & \best{0.1} & 0.0 & 0.0 & \best{0.9} & \best{0.6} & \best{0.6} \\

\bottomrule
\end{tabular}
\end{threeparttable}
}
\end{table*}

\subsection{Comparison with State-of-the-Art Methods}
\textbf{OpenOccupancy.}
Table~\ref{table_openoccupancy} reports the results on OpenOccupancy with v0.0 annotations under five sensor configurations. 
As no standard protocol exists for sensor outages in multimodal occupancy, we reproduce the competing methods under identical sensor configurations. With all sensors available, SpikeOcc remains competitive with strong ANN-based baselines~\citep{occmamba,spsc}, indicating that resilience is achieved without sacrificing standard occupancy performance. 
As sensor availability decreases, its advantage becomes progressively more evident, particularly under complete camera or LiDAR outages. The geometry trade-off is analyzed in \textit{Appendix~\ref{appg}}.

\textbf{SemanticKITTI.}
Table~\ref{table_semantikkiti_full} shows a consistent trend on SemanticKITTI. 
In this setting, only two camera views are available, making the model more dependent on LiDAR than on OpenOccupancy. 
SpikeOcc remains competitive under complete sensing and largely preserves its performance when all cameras are unavailable. 
More notably, its advantage becomes substantially larger under LiDAR outage, where existing multimodal models degrade severely~\citep{wang2024occgen}. 

\begin{wrapfigure}[13]{r}{0.345\columnwidth}
    \vspace{-29pt}
    \centering

    \includegraphics[
        width=\linewidth,
        height=0.2\textheight,
        keepaspectratio
    ]{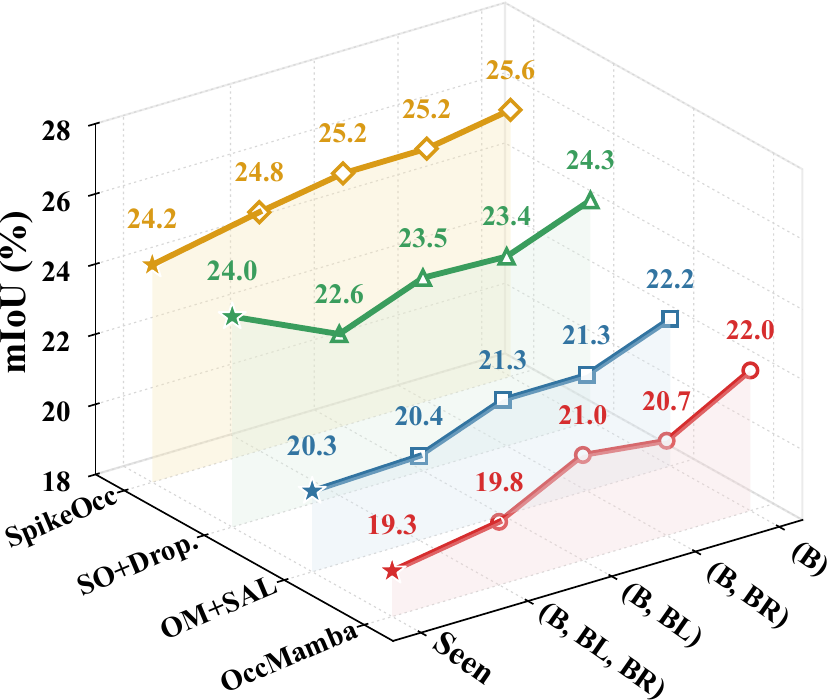}

    \caption{Generalization to unseen sensor configurations. B: Back, BL: Back left, BR: Back right.}
    \label{fig:unseen_configuration}

    \vspace{-8pt}
\end{wrapfigure}

\subsection{Ablation Study}
\textbf{Generalization to Unseen Sensor Configurations.}
Instead of memorizing a finite set of predefined failure patterns, a resilient occupancy model should generalize to unseen sensor configurations. We therefore evaluate all models on held-out sensor combinations, as shown in Fig.~\ref{fig:unseen_configuration}.
SpikeOcc maintains consistently strong performance across these unseen configurations, while SAL also improves the generalization of the ANN-based baseline~\citep{occmamba}.
Separately, we compare SAL with conventional Sensor Dropout~\citep{liu2017sensor},
which randomly removes sensor inputs during training to reduce dependence on any
specific sensor subset.
Beyond random sensor dropout, SAL turns sensor failures into a structured curriculum, yielding stronger generalization to unseen configurations.

\begin{table*}[t]
\centering

\begin{minipage}[t]{0.55\textwidth}
\vspace{0pt}
\centering

\captionof{table}{
\textbf{Ablation on SAL and MoE under different sensor settings.}
G and P denote GELU and PLIF respectively. M. means Missing. 
}
\label{table_abl_sal}

\scriptsize
\setlength{\tabcolsep}{4pt}
\renewcommand{\arraystretch}{0.92}

\newcommand{\abyes}{$\bullet$}
\newcommand{\abno}{$\circ$}
\newcommand{\abnone}{--}

\resizebox{\linewidth}{!}{%
\begin{tabular}{l|c|c|cc|cc|cc}
\toprule

\multirow{2}{*}{\textbf{Method}} &
\multirow{2}{*}{\textbf{SAL}} &
\multirow{2}{*}{\textbf{MoE}} &
\multicolumn{2}{c|}{\textbf{All Sensors}} &
\multicolumn{2}{c|}{\textbf{All Cam. M.}} &
\multicolumn{2}{c}{\textbf{LiDAR M.}}
\\

\cmidrule(lr){4-5}
\cmidrule(lr){6-7}
\cmidrule(lr){8-9}

& & &
\textbf{IoU} & \textbf{mIoU} &
\textbf{IoU} & \textbf{mIoU} &
\textbf{IoU} & \textbf{mIoU}
\\

\midrule

\multirow{4}{*}{OccMamba}
& \abno  & \abnone
& 35.7 & 26.2
& 18.8 & 9.4
& 7.8  & 1.6 \\

& \abyes & \abnone
& 36.2 & 26.4
& 29.3 & 16.1
& 19.2 & 8.5 \\

& \abno  & G
& 30.1 & 21.2
& 19.7 & 9.2
& 10.7 & 2.9 \\

& \abyes & G
& 33.3 & 23.4
& 29.9 & 13.7
& 16.3 & 6.8 \\

\midrule

\multirow{4}{*}{Naive SNN}
& \abno  & \abnone
& 35.9 & 16.7
& 21.8 & 8.7
& 0.3  & 0.0 \\

& \abyes & \abnone
& 31.2 & 22.4
& 29.3 & 16.1
& 18.8 & 8.9 \\

& \abno  & P
& 35.2 & 24.7
& 31.5 & 15.6
& 17.4 & 8.1 \\

&
\cellcolor{gray!12}\abyes &
\cellcolor{gray!12}P &
\cellcolor{gray!12}\textbf{35.8} &
\cellcolor{gray!12}\textbf{25.9} &
\cellcolor{gray!12}\textbf{36.8} &
\cellcolor{gray!12}\textbf{22.7} &
\cellcolor{gray!12}\textbf{19.7} &
\cellcolor{gray!12}\textbf{9.2} \\

\bottomrule
\end{tabular}%
}

\end{minipage}
\hfill
\begin{minipage}[t]{0.44\textwidth}
\vspace{0pt}
\centering

\captionof{table}{
\textbf{Parameter-matched comparison.}
Capacity scaling versus expert specialization.
All models are trained \textit{without SAL}.
}
\label{tab:param_matched}

\scriptsize
\setlength{\tabcolsep}{4.1pt}
\renewcommand{\arraystretch}{1.0}

\resizebox{\linewidth}{!}{%
\begin{tabular}{l|c|ccc}
\toprule

\textbf{Method} &
\textbf{Params} &
\textbf{Full} &
\makecell{\textbf{All Cam.}\\\textbf{M.}} &
\makecell{\textbf{LiDAR}\\\textbf{M.}}
\\

\midrule
OccMamba-128
& 64.6M
& 25.2
& 8.6
& 1.1 \\

OccMamba-384
& 109.2M
& 26.2
& 9.4
& 1.6 \\

OccMamba-512-ext
& 154.8M
& 26.4
& 9.5
& 2.9 \\

\midrule

Naive SNN
& 109.1M
& 16.7
& 8.7
& 0.0 \\

Dense SNN-S
& 111.1M
& 16.1
& 8.1
& 0.1 \\

Dense SNN-M
& 141.8M
& 18.0
& 8.5
& 1.6 \\

Dense SNN-L
& 158.1M
& 18.5
& 8.9
& 3.7 \\

\rowcolor{gray!12}
\textbf{SMoE}
& 112.3M
& \textbf{24.7}
& \textbf{15.6}
& \textbf{8.1} \\

\bottomrule
\end{tabular}%
}

\end{minipage}

\end{table*}

\begin{figure*}[t]
\centering

\begin{minipage}[t]{0.551\textwidth}
\vspace{0pt}
\centering

\includegraphics[
    width=\linewidth
]{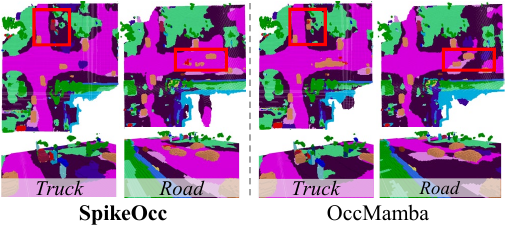}
\captionof{figure}{
\textbf{Qualitative comparison of occupancy predictions.} SpikeOcc better captures small objects and preserves the geometry of major scene structures.
}
\label{fig:qualitative}

\end{minipage}
\hfill
\begin{minipage}[t]{0.439\textwidth}
\vspace{0pt}
\centering

\includegraphics[
    width=\linewidth
]{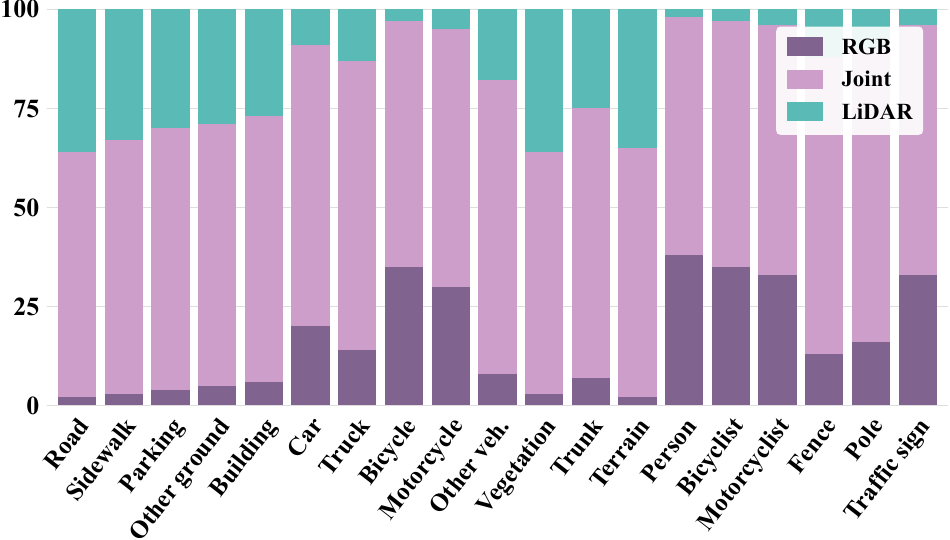}

\vspace{-0.35mm}
\captionof{figure}{Class-conditioned aggregation of scene-level routing weights on SemanticKITTI under complete sensing.}
\label{fig:expert_specialization}

\end{minipage}

\end{figure*}

\textbf{Ablation on SAL and MoE.} The variants differ only in the neuron operator, with identical architectures and training settings. Table~\ref{table_abl_sal} reveals two complementary effects. 
SAL consistently improves resilience across both ANN and SNN backbones, indicating that its benefit is not tied to a specific architecture. 
In contrast, while the ANN-based MoE brings limited benefit, the PLIF-based~\citep{fang2021plif} MoE substantially closes the semantic gap of the naive SNN across all sensor conditions. 
Combining SAL with spiking MoE yields the strongest overall resilience, while largely preserving the full-sensor performance. More SAL-equipped baselines are reported in \textit{Appendix~\ref{apph}}.

\textbf{Capacity Scaling.}
To isolate the effect of expert specialization, we construct parameter-matched dense SNNs by increasing network depth while keeping the feature dimension and operators unchanged~\citep{eigen2014learning,csordas2024moeut}. Table~\ref{tab:param_matched} shows simply scaling the dense architecture brings only modest improvements in mIoU, and even the Dense SNN-L remains clearly behind SMoE across all sensor settings. The gap suggests that expert specialization, rather than mere capacity scaling, is key to handling heterogeneous sensing dynamics.

\textbf{Expert Specialization.}
After training a spiking MoE, it is natural to study the usefulness of the learned router, in the light of several pitfalls~\citep{riquelme2021scaling}.
To this end, we analyze expert routing on SemanticKITTI and visualize the resulting occupancy predictions in Fig.~\ref{fig:qualitative} and different aggregated routing profiles in Fig.~\ref{fig:expert_specialization}. Expert weights remain shared across all voxels within each scene. The joint expert dominates most categories while RGB and LiDAR weights show distinct class-conditioned patterns.
These patterns suggest that the router adjusts the relative contributions of modality-specific and joint experts according to the semantic composition of each scene.

\textbf{Neuron Dynamics and Time Steps.}
We investigate neuron type and simulation length to understand how intrinsic spiking dynamics affect occupancy learning.
As shown in Table~\ref{tab:neuron_type}, PLIF outperforms conventional IF and LIF neurons~\citep{burkitt2006review}.
A plausible explanation is that adaptive membrane dynamics better capture heterogeneous multimodal features. Table~\ref{tab:temporal_config} further shows that additional time steps initially improve representation quality, but the gain saturates beyond $T=4$ while the firing rate continues to increase.

\textbf{Computational Efficiency.}
To characterize the practical cost of deployment, computational efficiency is evaluated in terms of inference latency, throughput, and theoretical energy consumption. All measurements are conducted on 8 NVIDIA A800 GPUs with batch size one under identical input settings after warm-up. For energy estimation, we follow the widely adopted 45-nm CMOS energy model~\citep{horowitz2014computing}, where a 32-bit MAC and AC operation consume 4.6 pJ and 0.9 pJ, respectively. The theoretical energy is calculated as $E_{\mathrm{theory}}=4.6N_{\mathrm{MAC}}+0.9N_{\mathrm{AC}}$, where $N_{\mathrm{MAC}}$ and $N_{\mathrm{AC}}$ are the operation counts in billions and $E_{\mathrm{theory}}$ is measured in mJ. Since the pretrained camera and LiDAR encoders are shared across all variants, MACs, ACs, and theoretical energy are reported for the fusion module only. Compared with M-CONet, SpikeOcc substantially reduces fusion computation and energy while maintaining comparable latency and throughput in Table~\ref{tab:efficiency}.

\begin{figure*}[t]
\centering
\includegraphics[width=\linewidth]{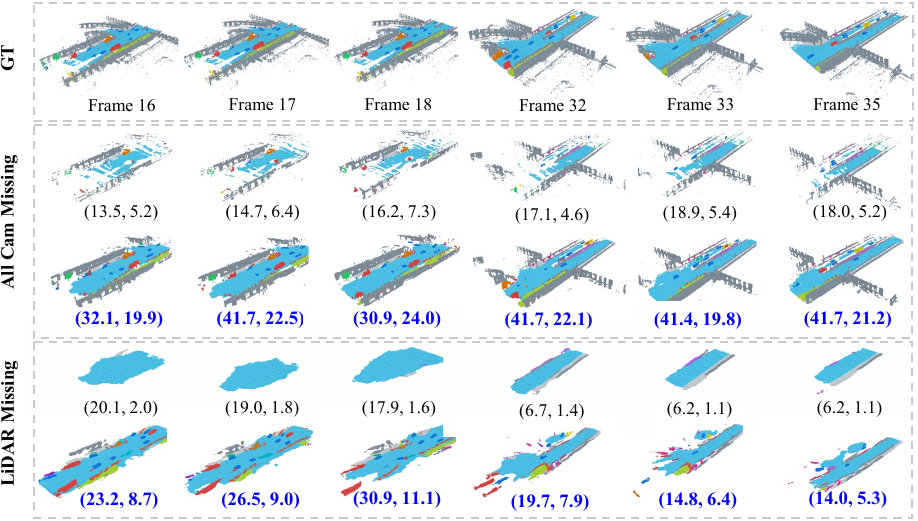} 
\caption{\textbf{More qualitative comparison with OccMamba.} SpikeOcc preserves more complete scene geometry and semantics under sensor outages. Numbers denote (IoU, mIoU).}
\label{fig:more_qualitative}
\end{figure*}

\textbf{Qualitative Analysis.}
Figures~\ref{fig:qualitative} and~\ref{fig:more_qualitative} provide qualitative evidence across SemanticKITTI and OpenOccupancy.
SpikeOcc better retains fine-grained objects, while its advantage becomes more pronounced under complete sensor outages, where competing predictions are often fragmented. 

\begin{table*}[t]
\centering

\makebox[\textwidth][c]{%
\begin{minipage}[t]{0.275\textwidth}
\vspace{0pt}
\centering

\captionof{table}{
Neuron type.
}
\label{tab:neuron_type}

\vspace{1pt}

\setlength{\tabcolsep}{3.0pt}
\renewcommand{\arraystretch}{1.05}

\resizebox{\linewidth}{!}{%
\begin{tabular}{lccc}
\toprule

\textbf{Neuron} &
$\boldsymbol{\tau}$ &
\makecell{\textbf{Learn.}\\$\boldsymbol{\tau}$} &
\textbf{mIoU} \\

\midrule

IF
& --
& --
& 25.2 \\

LIF
& 2.0
& $\times$
& 25.3 \\
\rowcolor{gray!12}
PLIF
& \textbf{2.0}
& $\checkmark$
& \textbf{25.9} \\

\bottomrule
\end{tabular}%
}

\end{minipage}%
\hfill
\begin{minipage}[t]{0.22\textwidth}
\vspace{0pt}
\centering

\captionof{table}{
Time steps.
}
\label{tab:temporal_config}

\vspace{1pt}

\setlength{\tabcolsep}{3.2pt}
\renewcommand{\arraystretch}{1.05}

\resizebox{\linewidth}{!}{%
\begin{tabular}{cccc}
\toprule

$\boldsymbol{T}$ &
$\boldsymbol{V_{\mathrm{th}}}$ &
\makecell{\textbf{Firing}\\\textbf{Rate (\%)}} &
\textbf{mIoU} \\

\midrule

1 & 0.5 & 7.6 & 24.0 \\
2 & 0.5 & 11.1 & 24.5 \\
\rowcolor{gray!12}
\textbf{4} & \textbf{0.5} & \textbf{12.7} & \textbf{25.9} \\
6 & 0.5 & 13.4 & 25.7 \\

\bottomrule
\end{tabular}%
}

\end{minipage}%
\hfill
\begin{minipage}[t]{0.48\textwidth}
\vspace{0pt}
\centering

\captionof{table}{
Computational efficiency.
}
\label{tab:efficiency}

\vspace{1pt}

\setlength{\tabcolsep}{2.3pt}
\renewcommand{\arraystretch}{1.05}

\resizebox{\linewidth}{!}{%
\begin{tabular}{lccccc}
\toprule

\textbf{Variant} &
\makecell{\textbf{Lat.}\\\textbf{(ms)}} &
\textbf{FPS} &
\makecell{\textbf{MACs}\\\textbf{(G)}} &
\makecell{\textbf{ACs}\\\textbf{(G)}} &
\makecell{\textbf{Energy}\\\textbf{(mJ)}} \\

\midrule

M-CONet
& 1918.8 & 4.17 & 14.7 & 0 & 67.6 \\
SpikeOcc-LIF
& 1862.0 & 4.30 & 1.68 & 1.63 & 9.2 \\
\rowcolor{gray!12}
SpikeOcc-PLIF
& 1797.3 & 4.45 & 1.68 & 1.86 & 9.4 \\

\bottomrule
\end{tabular}%
}

\end{minipage}%
}

\end{table*}

\section{Conclusion}
We present SpikeOcc, a spike-driven framework for resilient multimodal 3D occupancy prediction, targeting sensor failures that are inevitable in real-world perception systems.
Rather than relying on depth scaling, a width-first spiking Mixture-of-Experts restructures multimodal fusion into specialized pathways, enabling adaptive fusion across changing sensing regimes.
Complementing this, Sensor Availability Learning organizes diverse sensor states into a staged curriculum, gradually strengthening the model against increasingly incomplete observations.
Extensive experiments on OpenOccupancy and SemanticKITTI demonstrate strong performance under both complete sensing and challenging modality outages.
By supporting flexible inference across varying sensor availability, SpikeOcc advances occupancy prediction toward resilient deployment and offers a fresh perspective for future multimodal perception research. Limitations are discussed in \textit{Appendix~\ref{appi}}.

\subsection*{AI use statement}
In this work, we used generative AI tools to aid or polish writing. We take responsibility for the final content of this work,
including text, claims or artifacts produced with the aid of generative AI.
\subsection*{Ethics statement}
This work uses publicly available datasets and does not involve human-subject
experiments or the collection of private personal data.
We are not aware of any conflicts of interest or additional ethical concerns
associated with this work.
\subsection*{Reproducibility statement}
We provide the model formulation, training details, evaluation protocol, and
additional implementation details in the main paper and appendix.
All experiments follow the official dataset splits and are conducted with
three random seeds.
Anonymous source code, configuration files, and training scripts are provided
through the OpenReview Supplementary Material.



\bibliography{iclr2027_conference}
\bibliographystyle{iclr2027_conference}

\appendix
\section{Why Spiking Representations for Incomplete Sensing?}\label{appa}

We provide a representation-level interpretation of why discrete spiking
features can be attractive under varying sensor availability.
The analysis concerns the binary spike codes generated inside the spiking
experts, rather than claiming that an arbitrary SNN is intrinsically robust to arbitrary sensor failures. In particular, we show that hard-threshold
spiking introduces an explicit margin within which sensor-induced feature
perturbations leave the discrete representation exactly unchanged.
\paragraph{Sensor-induced perturbation.}
Consider the pre-threshold membrane potential of a spiking layer at one
internal time step, denoted by
$\mathbf{u}\in\mathbb{R}^{d}$ under complete sensing.
For notational simplicity, the time index is omitted; the following argument
applies independently to every internal step.
Let $M$ denote a sensor-availability state.
The corresponding membrane potential can always be written as
\begin{equation}
    \mathbf{u}^{M}
    =
    \mathbf{u}
    +
    \boldsymbol{\Delta}^{M},
    \label{eq:app_sensor_perturbation}
\end{equation}
where $\boldsymbol{\Delta}^{M}$ represents the representation perturbation
induced by changing sensor availability.
The binary spike codes before and after this perturbation are
\begin{equation}
    s_i
    =
    \mathbb{H}(u_i-V_{\mathrm{th}}),
    \qquad
    s_i^{M}
    =
    \mathbb{H}(u_i+\Delta_i^{M}-V_{\mathrm{th}}),
    \label{eq:app_spike_codes}
\end{equation}
where $\mathbb{H}$ is the hard-threshold function used in Eq.~\ref{eq5}.
We define the threshold margin of the $i$-th unit as
\begin{equation}
    \gamma_i
    =
    \left|u_i-V_{\mathrm{th}}\right|.
    \label{eq:app_margin}
\end{equation}

\noindent\textbf{Proposition 1 (Threshold-margin stability).}
If the spike state of unit $i$ changes under sensor state $M$, then the
sensor-induced perturbation must be at least as large as its threshold
margin:
\begin{equation}
    s_i^{M}\neq s_i
    \quad\Longrightarrow\quad
    \left|\Delta_i^{M}\right|
    \ge
    \gamma_i .
    \label{eq:app_margin_bound}
\end{equation}
Equivalently,
\begin{equation}
    \mathbbm{1}[s_i^{M}\neq s_i]
    \le
    \mathbbm{1}
    \left[
        \left|\Delta_i^{M}\right|
        \ge
        \gamma_i
    \right].
    \label{eq:app_flip_indicator}
\end{equation}

\noindent\textit{Proof.}
If $u_i\ge V_{\mathrm{th}}$ but
$u_i+\Delta_i^{M}<V_{\mathrm{th}}$, then
$|\Delta_i^{M}|\ge u_i-V_{\mathrm{th}}=\gamma_i$.
Conversely, if $u_i<V_{\mathrm{th}}$ but
$u_i+\Delta_i^{M}\ge V_{\mathrm{th}}$, then
$|\Delta_i^{M}|\ge V_{\mathrm{th}}-u_i=\gamma_i$.
These are the only two cases in which the binary spike state can change,
which proves Eq.~\ref{eq:app_margin_bound}.
\hfill$\square$

Proposition~1 gives an explicit invariance region for the forward spike
representation. In particular, if
$|\Delta_i^{M}|<\gamma_i$, the spike state is exactly preserved even though
the underlying continuous membrane potential has changed.
For the complete spike vector, let
$d_H(\mathbf{s},\mathbf{s}^{M})$ denote the Hamming distance.
Eq.~\ref{eq:app_flip_indicator} immediately yields
\begin{equation}
    d_H(\mathbf{s},\mathbf{s}^{M})
    \le
    \sum_{i=1}^{d}
    \mathbbm{1}
    \left[
        |\Delta_i^{M}|
        \ge
        \gamma_i
    \right].
    \label{eq:app_hamming_bound}
\end{equation}
Hence, sensor-induced feature changes do not necessarily propagate to all
spiking units; only perturbations that cross their corresponding firing
margins can alter the discrete code.

This property differs from a locally smooth continuous representation.
For a differentiable continuous mapping
$\mathbf{a}=g_{\mathrm{cont}}(\mathbf{u})$, a sufficiently small perturbation satisfies
the first-order expansion
\begin{equation}
    g_{\mathrm{cont}}(\mathbf{u}+\boldsymbol{\Delta}^{M})
    -
    g_{\mathrm{cont}}(\mathbf{u})
    =
    \mathbf{J}_{g_{\mathrm{cont}}}(\mathbf{u})
    \boldsymbol{\Delta}^{M}
    +
    o\!\left(
        \|\boldsymbol{\Delta}^{M}\|
    \right),
    \label{eq:app_ann_local}
\end{equation}
where $\mathbf{J}_{g_{\mathrm{cont}}}$ denotes the local Jacobian.
Whenever this Jacobian is nonzero along the perturbation direction,
arbitrarily small input changes generally induce nonzero feature changes.
In contrast, hard-threshold spiking provides a finite neighborhood in which
the discrete forward representation remains exactly invariant.
\section{Diagnosing Instability in Naive Spiking Fusion}\label{appb}
\begin{figure*}[t]
    \centering

    \begin{minipage}[t]{0.44\textwidth}
        \vspace{0pt}
        \centering

        \captionof{table}{
        \textbf{Effect of curriculum ordering in SAL.}
        All variants use the same sensor states, while differing only in their
        presentation order. Results are reported in mIoU.
        }
        \label{tab:sal_order}

        \small
        \setlength{\tabcolsep}{4.5pt}
        \renewcommand{\arraystretch}{1.05}

        \begin{tabular}{l|ccc}
            \toprule
            Schedule &
            Full &
            All Cam. M. &
            LiDAR M. \\
            \midrule
            Static
            & 25.4 & 21.5 & 8.5 \\
            Shuffled
            & 25.2 & 21.0 & 8.2 \\
            Reverse
            & 24.8 & 20.2 & 7.7 \\
            \rowcolor{gray!12}
            \textbf{SAL}
            & \textbf{25.9} & \textbf{22.7} & \textbf{9.2} \\
            \bottomrule
        \end{tabular}

    \end{minipage}
    \hfill
    \begin{minipage}[t]{0.53\textwidth}
        \vspace{0pt}
        \centering

        \includegraphics[width=\linewidth]{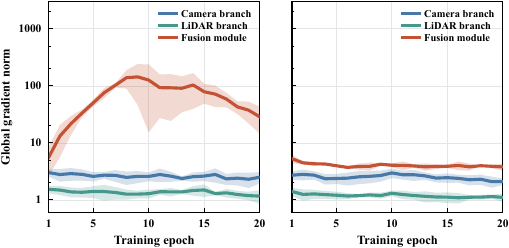}

        \captionof{figure}{
        \textbf{Branch-wise gradient during training.}
        }
        \label{fig:branch_gradient}

    \end{minipage}

\end{figure*}

To further diagnose the instability observed in Fig.~\ref{fig2}, we analyze the
branch-wise gradient dynamics of the camera branch, LiDAR branch, and fusion
module. As shown in Fig.~\ref{fig:branch_gradient}, the camera and LiDAR branches of the Naive
SNN remain relatively stable, while the fusion module exhibits much larger
gradient magnitudes and repeated fluctuations. In contrast, SpikeOcc maintains much smoother gradient
dynamics across all branches. In particular, the large oscillations in the
fusion module are substantially reduced, indicating that the proposed expert
decomposition alleviates interference between heterogeneous modalities and
stabilizes spiking multimodal fusion.

\section{Alternative Curriculum Orderings}\label{appc}
We examine whether the ordering of sensor failures matters in SAL.
All variants use the same sensor states and comparable overall exposure,
while only the training schedule is changed.
As shown in Table~\ref{tab:sal_order}, the default curriculum consistently
outperforms alternative orderings, particularly under complete modality
outages. Reversing the curriculum leads to the largest degradation, suggesting
that exposing the model to severe failures too early hinders the formation of
stable multimodal representations. Static and shuffled schedules perform
better than the reversed order, but remain inferior to the progressive
curriculum. These results indicate that SAL benefits not only from exposure to
sensor failures, but also from gradually increasing their severity during
training.

\section{Routing Target Definition}\label{appd}
The routing target in Eq.~\ref{eq:route_loss} is determined by the
availability of the two modalities. Let $c$, $l$, and $j$ denote the RGB,
LiDAR, and joint experts, respectively. For a sensor state characterized by
$m_c$ and $m_l$, we define
\begin{equation}
y^{\mathrm{route}} =
\begin{cases}
c, & m_c=1,\; m_l=0,\\
l, & m_c=0,\; m_l=1,\\
j, & m_c=1,\; m_l=1.
\end{cases}
\label{eq:route_target}
\end{equation}
Thus, LiDAR missing states are assigned to the RGB expert, while complete
camera outages are assigned to the LiDAR expert. When both modalities remain
available, including partial camera failures, the joint expert is used as the
routing target. These cases cover all valid sensor states because
$m_c+m_l\geq 1$. The target is used only by the auxiliary cross entropy loss on the raw router
logits. The final representation still follows the soft routing weights in
Eq.~\ref{eq:router_weights}. Therefore, the auxiliary target encourages expert
specialization without enforcing hard expert selection.

\section{Additional Evaluation Details}
\label{appe}

We follow the standard occupancy evaluation protocol and report geometric
IoU and semantic mIoU~\citep{openoccupancy}. For geometric occupancy, all non-empty semantic classes are merged into the
occupied category. The IoU is computed as
\begin{equation}
\mathrm{IoU}
=
\frac{\mathrm{TP}}
{\mathrm{TP}+\mathrm{FP}+\mathrm{FN}},
\label{eq:geo_iou}
\end{equation}
where TP, FP, and FN denote the numbers of true-positive, false-positive,
and false-negative occupied voxels, respectively.

For semantic occupancy, the IoU of class $k$ is
\begin{equation}
\mathrm{IoU}_k
=
\frac{\mathrm{TP}_k}
{\mathrm{TP}_k+\mathrm{FP}_k+\mathrm{FN}_k},
\label{eq:sem_iou}
\end{equation}
and the mean IoU is
\begin{equation}
\mathrm{mIoU}
=
\frac{1}{K-1}
\sum_{k=1}^{K-1}\mathrm{IoU}_k,
\label{eq:miou}
\end{equation}
where class $0$ denotes free space and is excluded from the semantic average.
Thus, IoU measures geometric occupancy quality, while mIoU evaluates
class-aware prediction over occupied voxels.
\section{Inference Protocol}
\label{appf}
During training, SAL samples different sensor states according to the
curriculum schedule, and the auxiliary routing loss in
Eq.~\ref{eq:route_loss} is used to encourage expert specialization.
The corresponding sensor masks are propagated through the camera and LiDAR
branches to construct the masked inputs.

At inference time, no curriculum sampling or routing supervision is used.
The model directly receives the currently available sensor inputs together
with their availability descriptor, and performs a single forward pass.
The router produces the soft expert weights defined in
Eq.~\ref{eq:router_weights}, after which the fused representation is sent to
the occupancy head. No retraining, test-time optimization, or repeated
inference is required for different sensor configurations.

\section{Geometry--Semantics Trade-off under Camera Outages}
\label{appg}

An interesting trend appears as camera availability decreases in Table~\ref{table_openoccupancy}. The geometric
IoU of SpikeOcc slightly increases, while the semantic mIoU gradually drops.
A plausible explanation is that LiDAR provides direct and reliable geometric
measurements, whereas camera features contribute richer appearance cues for
semantic discrimination. Similar modality complementarity has also been
observed in multimodal occupancy methods~\citep{pan2024coocc}. As fewer cameras
remain available, the prediction becomes more dependent on LiDAR features,
which can preserve or even simplify geometric occupancy estimation but provides
less semantic information. This observation suggests that geometric robustness
and semantic richness are not always aligned under incomplete sensing, further
motivating the preservation of modality-specific strengths in multimodal
fusion.
\section{SAL-Equipped Baseline Comparisons}
\label{apph}
\newcommand{\gain}[1]{%
\,\raisebox{-0.45ex}{\scriptsize\textcolor{green!50!black}{(+#1)}}%
}
\newcommand{\drop}[1]{%
\,\raisebox{-0.45ex}{\scriptsize\textcolor{red!65!black}{(-#1)}}%
}

\begin{table}[t]
\centering
\caption{
\textbf{Comparison with SAL-equipped baselines on OpenOccupancy.}
Values in parentheses denote the change relative to the corresponding
model trained without SAL.
}
\label{tab:sal_baselines}

\small
\setlength{\tabcolsep}{5pt}
\renewcommand{\arraystretch}{1.08}

\begin{tabular}{l|cc|cc|cc}
\toprule
\multirow{2}{*}{Method} &
\multicolumn{2}{c|}{Full} &
\multicolumn{2}{c|}{All Cam. Missing} &
\multicolumn{2}{c}{LiDAR Missing} \\
& IoU & mIoU & IoU & mIoU & IoU & mIoU \\
\midrule

OccMamba + SAL
& \textbf{36.2}\gain{0.5}
& \textbf{26.4}\gain{0.2}
& 29.3\gain{10.5}
& 16.1\gain{6.7}
& 19.2\gain{11.4}
& 8.5\gain{6.9} \\

SPSC + SAL
& 35.4\gain{0.2}
& 26.2\gain{0.1}
& 26.7\gain{4.9}
& 13.9\gain{5.2}
& 16.5\gain{16.3}
& 6.0\gain{5.9} \\

\midrule

\rowcolor{gray!12}
\textbf{SpikeOcc}
& 35.8\gain{0.6}
& 25.9\gain{1.2}
& \textbf{36.8}\gain{5.3}
& \textbf{22.7}\gain{7.1}
& \textbf{19.7}\gain{2.3}
& \textbf{9.2}\gain{1.1} \\

\bottomrule
\end{tabular}
\end{table}

To separate the contribution of SAL from the proposed architecture, we apply
the same SAL schedule and sensor masks to representative multimodal occupancy
baselines. Table~\ref{tab:sal_baselines} reports the results under matched
availability training.
SAL consistently improves the resilience of different multimodal architectures,
confirming that its benefit is not specific to SpikeOcc. Nevertheless, under the same
availability training, SpikeOcc retains a clear advantage in severe sensor
outages. In particular, it reaches 22.7 mIoU with all cameras missing, compared
with 16.1 for OccMamba with SAL. These results indicate that SAL provides a
general training benefit, while the additional robustness of SpikeOcc arises
from its spiking expert architecture.

\section{Limitations and Future Directions}
\label{appi}

SpikeOcc focuses on a controlled setting in which sensor availability changes are known to the model. Within this scope, SMoE and SAL provide a unified predictor that avoids retraining separate models for each sensor configuration and substantially improves robustness under severe outages. Several limitations, however, remain beyond the current formulation.

First, the availability descriptor assumes that the system knows which sensors are currently valid. In practice, a sensor may produce corrupted observations before a complete failure is detected, making sensor health itself uncertain. Jointly estimating sensor reliability would require an additional fault detection or uncertainty estimation mechanism, which introduces a different learning problem from the controlled missing sensor setting studied here. Future work can therefore extend the router with learned sensor health estimation when computing expert weights.

Second, the space of possible sensor failures grows rapidly with the number of sensors. Even with six cameras and one LiDAR, there are many possible nonempty availability combinations, while real deployment further includes continuous degradations such as partial occlusion, calibration errors, and corrupted measurements. Exhaustively including these cases during training is neither practical nor desirable, since it would obscure the controlled analysis of representative failures. SAL instead uses a small set of structured states and evaluates whether the learned model generalizes beyond them. A natural extension is to replace the predefined curriculum with learned or adaptive failure sampling, allowing difficult configurations and continuous sensor degradation to be introduced according to the current model state.

Finally, SpikeOcc is designed as a single multimodal model rather than a collection of modality specific specialists. Consequently, when only one modality remains available, a dedicated unimodal model trained exclusively for that input can still retain advantages in some settings. Our objective is different. The goal is to preserve a strong performance floor without switching models or retraining when the sensing configuration changes. Future work may combine the current shared framework with stronger modality specific priors or lightweight specialist adaptation while retaining a unified inference pipeline.
\end{document}